\documentclass[letterpaper, 10 pt, conference]{ieeeconf}  

\IEEEoverridecommandlockouts                              
\usepackage[utf8]{inputenc}
\usepackage[T1]{fontenc}
\usepackage{mathptmx} 
\usepackage{mathptmx} 
\usepackage{amsmath} 
\usepackage{amssymb}  
\usepackage{graphicx} 
\usepackage{grffile}
\usepackage{silence}  
\usepackage[hypcap=true]{caption}
\usepackage{array}
\usepackage{xcolor}
\usepackage{mathtools}
\usepackage{commath}
\usepackage{colortbl}
\usepackage{enumerate}
\usepackage{hyperref}
\usepackage{pifont}

\usepackage{float}
\usepackage{tabularx}
\usepackage{makecell}
\usepackage{multirow}
\usepackage{placeins}
\usepackage{hyperref}
\usepackage{tabu}
\usepackage{cite} 
\usepackage{balance}
\usepackage{tikz}
\usepackage{pgfplots}
\usepackage{subfig}

\title{\LARGE \bf
Fall Prediction for Bipedal Robots: The Standing Phase
}

\author{ M. Eva Mungai, Gokul Prabhakaran, and Jessy W. Grizzle %
\thanks{\{Mechanical Engineering, Robotics, Robotics\} Department, University of Michigan, Ann Arbor, MI, USA.{\text{\{mungam,gok,grizzle\}@umich.edu}}}
}

\begin{document}
\maketitle
\thispagestyle{empty}

\begin{abstract}
This paper presents a novel approach to fall prediction for bipedal robots, specifically targeting the detection of potential falls while standing caused by abrupt, incipient, and intermittent faults. Leveraging a 1D convolutional neural network (CNN), our method aims to maximize lead time for fall prediction while minimizing false positive rates. The proposed algorithm uniquely integrates the detection of various fault types and estimates the lead time for potential falls. Our contributions include the development of an algorithm capable of detecting abrupt, incipient, and intermittent faults in full-sized robots, its implementation using both simulation and hardware data for a humanoid robot, and a method for estimating lead time. Evaluation metrics, including false positive rate, lead time, and response time, demonstrate the efficacy of our approach. Particularly, our model achieves impressive lead times and response times across different fault scenarios with a false positive rate of 0. The findings of this study hold significant implications for enhancing the safety and reliability of bipedal robotic systems. 
\end{abstract}


\section{Introduction}
The distinct morphology of bipedal robots endows them with the unique capability to navigate diverse terrains, from unstructured environments to human-centric spaces, making them ideal candidates for assisting in daily tasks and critical situations. Nevertheless, the real-world deployment of bipedal robots remains limited. Their high-dimensional, hybrid nature, combined with occasional stringent constraints, complicates the achievement of stable motion, particularly when confronted with disturbances.

\subsection{Motivation}
Although a well-designed controller can counteract certain disturbances, it is infeasible to anticipate every potential disruption a bipedal robot might face in real-world scenarios. This unpredictability underscores the inevitability of falls. To truly harness the capabilities of bipedal robots, there is a pressing need for fall prediction algorithms that can foresee a fall with ample lead time, where lead time denotes the interval between the fall prediction and its actual occurrence. Addressing this need, our paper presents a 1D convolutional neural network-based fall prediction algorithm. This algorithm not only predicts falls but also estimates the lead time, specifically for the standing task with the Digit bipedal robot \cite{ref:ar_digit}. \textit{We focus on the standing task as it offers a simplified context for fall prediction while laying the groundwork for more dynamic movements.}

\subsection{Background}
Falls in bipedal robots can often be traced back to faults, which are characterized as unforeseen deviations in one or more operational variables. Faults can occur from failures in the robot's hardware or software components, or be triggered by external factors. Irrespective of their origin, faults can be classified based on their temporal behavior into three categories: abrupt, incipient, and intermittent. Abrupt faults manifest as sudden or rapid changes, incipient faults evolve gradually over time, and intermittent faults appear sporadically \cite{ref:ISERMANN19827}. Each of these fault types can arise during real-world operations. For example, abrupt faults might be triggered by unexpected interactions with the environment (e.g., stepping in a hole), incipient faults could stem from discrepancies in the model (e.g., poor trajectory tracking in the operational space of the robot), and intermittent faults might emerge in unpredictable environments laden with obstacles. In this context, we term faults that precipitate a fall as \textbf{critical faults} and the paths leading to a fall as \textbf{unsafe trajectories}.

\begin{figure}[t]
    \centering
    \includegraphics[scale=0.45]{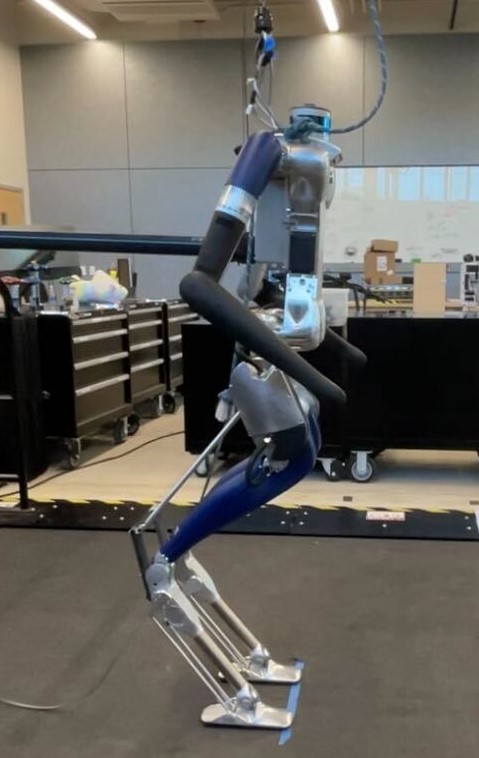}
    \caption{The experimental setup that was used to collect hardware data with the Digit robot \cite{ref:ar_digit}.}
    \label{fig:experimental_setUp}
\end{figure}

\subsection{Literature Review}
Existing fall prediction algorithms for bipedal robots predominantly target the detection of critical abrupt faults. The primary method consists of establishing a threshold that distinguishes these faults from regular data. Selecting a threshold is a delicate task due to the impracticality of accounting for every potential critical fault and the complexity of capturing (e.g., in a model) all the robot's safe states. Additionally, the presence of faulty states in the data is infrequent.

The overarching aim of fall prediction algorithms is to extend the lead time while curtailing the false positive rate. Consequently, their performance is typically assessed based on these two metrics \cite{ref:subburaman2023survey}. On the one hand, it's important to note that there's an inherent positive correlation between lead time and false positive rate. On the other hand, the impact of false negatives can be alleviated by setting thresholds on kinematic signals, such as the height of the center of mass, as elaborated in \cite{ref:kalyanakrishnan2011learning}. The minimum lead time deemed acceptable varies depending on the specific robot and the selected recovery algorithm.

In the realm of fall prediction, thresholds are derived from various sources: analytical models like \cite{ref:muender, ref:li_model_based, ref:stephens, ref:jalgha, ref:mummolo2017numerical}, hand-crafted features as seen in \cite{ref:ruiz2010fall, ref:hohn2006detection, ref:subburaman, ref:tay2016fall, ref:tran}, or data-driven models such as \cite{ref:kalyanakrishnan2011learning, ref:wu2021falling, ref:kormushev2012anatomy, ref:karssen2009fall, ref:suetani2011nonlinear, ref:marcolino2013detecting, ref:kim, ref:andre, ref:liu}. While analytical models offer a structured approach, they can be hampered by model uncertainties and might not fully encapsulate the robot's full dynamics, especially if based on simplifying assumptions. Simple thresholds are elusive for multifaceted systems like bipedal robots. On the other hand, while data-driven models might necessitate extensive data and remain constrained to the data's distribution, their popularity is on the rise, thanks to advancements in machine learning and computational capabilities. While both shallow methods \cite{ref:kalyanakrishnan2011learning, ref:wu2021falling, ref:kormushev2012anatomy, ref:karssen2009fall, ref:suetani2011nonlinear, ref:marcolino2013detecting, ref:kim} and deeper neural network-based approaches \cite{ref:andre, ref:liu} have found their place in the literature, fall prediction algorithms implemented for full-sized robots are generally based on shallow methods \cite{ref:kalyanakrishnan2011learning, ref:wu2021falling, ref:subburaman2023survey, ref:marcolino2013detecting}.

\subsection{Objectives of the Paper}
Our goals are as follows:
\begin{enumerate}
    \item Detect imminent falls caused by abrupt, incipient, and intermittent faults, ensuring an adequate lead time for the task of standing with a full-sized robot.
    \item Optimize the trade-off between maximizing lead time and minimizing false positive rates.
    \item Accurately estimate the lead time.
\end{enumerate}
The robot is deemed to have fallen if the height of its center of mass is less than 10 percent of its initial height. Drawing parallels with our prior research \cite{ref:mungai2023optimizing}, we adopt 0.2s as the minimum desired lead time, aligning with the requirements of reflexive algorithms such as those in \cite{ref:hohn2009probabilistic, ref:wu2021falling}. Achieving this objective presents challenges, given the controller's masking effects, the crowding phenomenon induced by incipient faults \cite{ref:safaeipour2021survey}, the direct correlation between false positive rates and lead time, the sporadic nature of intermittent faults, and the diminishing number of data points with increasing lead time.

To address these challenges, we propose a 1D convolutional neural network (CNN)-driven fall prediction algorithm, enhanced with several components. Our choice of a deep model is motivated by the significant impact of user-selected features on lead time and false positive rates when using shallow methods, as evidenced in our earlier work \cite{ref:mungai2023optimizing}. Furthermore, extracting these features for complicated bipedal robots, like Digit, is non-trivial. Our choice of a 1D CNN is underpinned by its equivariance to translation and proficiency in discerning local patterns in data with a grid structure. It is noteworthy that a bipedal robot's trajectory can be mapped onto a 1D grid, sampled at consistent (uniform) intervals \cite{ref:goodfellow2016deep}.

\subsection{Contributions}
We present several key contributions in this work:
\begin{itemize}
    \item Introduction of an algorithm capable of detecting abrupt, incipient, and intermittent faults in full-sized robots undertaking a standing task.
    \item Successful implementation of our fall prediction algorithm, both in simulation and on hardware, tailored for a full-sized humanoid robot.
    \item Development of an accurate method to estimate lead time.
    \item Release of an open-source dataset of a full-sized bipedal robot comprised of simulation and hardware trajectories with various critical and non-critical faults. The dataset can be accessed at \cite{ref:digit_dataset}.
\end{itemize}

\begin{figure}
\vspace{2mm}
    \centering
    \includegraphics[scale=0.55]{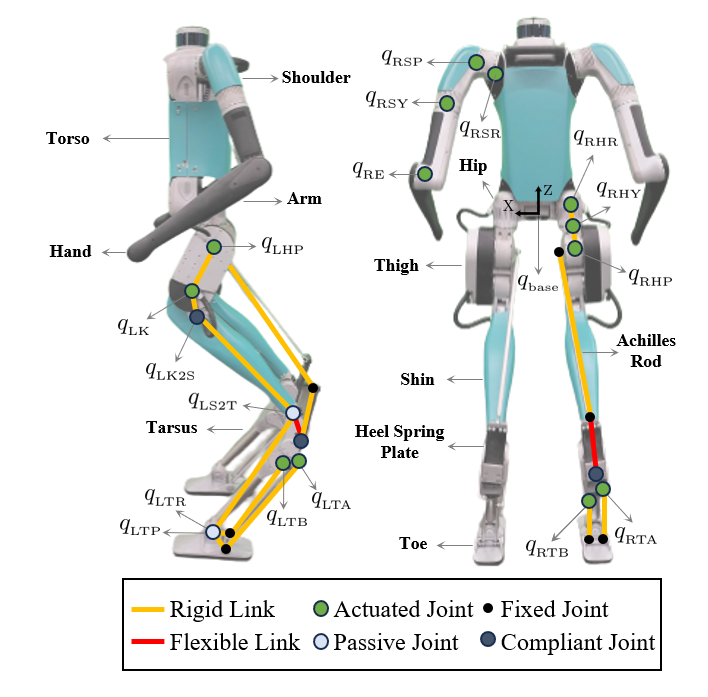}
    \caption{Kinematics architecture of the Digit robot by Agility Robotics \cite{ref:ar_digit}. Image Credit: Grant Gibson \cite{ref:grant}.}
    \label{fig:digitHardware}
    \vspace{-2mm}
\end{figure}

\section{Hardware Overview and Data Generation}
This section provides an overview of the bipedal robot, Digit, and outlines our approach to data generation. 

\subsection{Hardware Overview}
Developed by Agility Robotics, Digit is a state-of-the-art bipedal robot \cite{ref:ar_digit}. While it draws inspiration from Agility Robotic's earlier model, Cassie, Digit distinguishes itself with the addition of a torso and an integrated perception system. Possessing 30 degrees of freedom, Digit has 20 actuated joints. Weighing in at 48kg, its lower limb design is inspired by a Cassowary bird, leading to the unique nomenclature where what would typically be termed ``feet'' are actually ``toes''; we will use the latter terminology. The kinematic architecture of Digit is illustrated in Figure \ref{fig:digitHardware}.

\subsection{Simulation Data Generation}
\label{sec:sim_data}
We employ Agility's MuJoCo-based simulator in conjunction with a standing controller. This controller is designed to maintain both the center of mass and the zero moment point within the support polygon \cite{ref:grant}. We generate 900 trajectories each for abrupt and incipient faults, and 100 trajectories for intermittent faults. These faults are simulated by applying forces of various magnitudes to the robot's torso in the $x$-direction (i.e., sagittal plane).

Abrupt faults are simulated using impulsive forces with a duration of 0.075s, randomly uniformly distributed within a range of 0 - 414.8N. In the case of incipient faults, their crowding effect is captured through trapezoidal force profiles \cite{ref:he2018incipient}. 
These profiles have a slope of $\frac{480N}{s}$ over a varying duration to result in a desired constant amplitude over a time duration of 1s; the resulting force amplitudes of incipient faults are randomly uniformly distributed between 0 - 57.6N. Similar to \cite{ref:kalyanakrishnan2011learning} and \cite{ref:mungai2023optimizing}, the force ranges for both abrupt and incipient faults are calibrated to ensure an equal distribution of falling and safe trajectories. Emulating the unpredictable nature of intermittent faults, we apply two distinct forces. These forces are designed to mimic either abrupt or incipient faults. The first force's magnitude remains within the safe range, while the second force's magnitude can potentially lead to a fall or maintain stability.

To simulate minor disturbances that might induce slight oscillations in the robot's standing posture, we introduce a random impulsive force with a 0.075s duration, ranging from 0 - 202.4N, at the start of each trajectory. The abrupt and incipient faults are subsequently introduced between 2 - 3.5s following this oscillatory perturbation for all three faults. The time between the application periods for the two faults comprising the intermittent fault is 2s.
\subsection{Hardware Data Generation}
\label{sec:fdd_digit_hardware_generation}
To prevent the Digit robot from getting damaged during data collection, the hardware data generation is carried out with Digit attached to a gantry via a slack cable that allows Digit to move about. Additionally, the motor power is ``killed'' when the robot starts to fall, thereby allowing the gantry to catch it. Impulsive and trapezoidal forces are introduced to the robot's torso by pushing Digit with a pole as depicted in Figure \ref{fig:experimental_setUp}. To emulate the trapezoidal forces that result in an incipient fault, the pole is first rested on Digit before pushing. 
The approximate time of force application is obtained by coordinating the push on the robot with a keyboard press and evaluating the center of mass data. 

Twenty-seven (27) safe and 13 unsafe trajectories are collected for abrupt faults, while 26 safe and 15 unsafe trajectories are collected for incipient faults. Out of the abrupt and incipient unsafe trajectories, six and ten trajectories, respectively, exhibited trajectory profiles resembling those observed in simulations characterized by lower falling forces. The remaining falling trajectories exhibited similarities to simulation profiles featuring forces within the mid-range of the falling forces. Figure \ref{fig:experimental_setUp} depicts the experimental setup of the hardware data.

\subsection{Data Pre-processing}
\label{sec:pre-processinng}
From the 1,800 simulation-generated trajectories for abrupt and incipient faults, 200 are reserved for testing. The remaining trajectories are divided into training (80\%) and validation sets. The testing set is further supplemented with intermittent fault data and hardware data. We employ scikit-learn's stratified train-test split method for segmenting the simulation data and its min-max scaler to normalize the data within the range  $[0,1]$. It's important to note that only transient data is utilized during training. 

The trajectories have a sampling rate of approximately 40Hz, and sliding windows containing 10 data points are utilized to prioritize the most recent data points. To ensure that the robot oscillates slightly even in the presence of perturbing forces towards the high end of the range described in Section \ref{sec:sim_data}, only windows during and after the fault introduction are retained for the simulation data. For both hardware and simulation data, the features are transformed to the world coordinate located on the ground between the toes. Lastly, to adjust for any drift in the hardware data, the initial feature values are subtracted from subsequent data points.

\section{Fall Prediction Method}
The task of detecting critical faults and estimating lead time can be framed as a regression problem where lead time serves as the predicted variable. Given our definitions of lead time and critical faults, data points from unsafe trajectories prior to a critical fault's onset, along with all data points from safe trajectories, are assigned an infinite lead time. Data points following the introduction of a critical fault have a lead time within the range $[0, H]$, where $H>0$ represents the \textbf{maximum prediction horizon} (i.e, maximum interval of time over which fault prediction is attempted). Predicting a lead time less than $H$ can thus indicate the presence of critical faults. However, as illustrated in Figure \ref{fig:data_points_vs_windows_data}, the quantity of data points diminishes exponentially with increasing lead time, leading to an imbalanced regression problem \cite{ref:krawczyk2016learning,ref:yang2021delving, ref:torgo2015resampling, ref:branco2017smogn}.

\begin{figure}
    \vspace{2mm}
    \centering
\begin{tikzpicture}[scale=0.7]

\definecolor{darkgray176}{RGB}{176,176,176}
\definecolor{steelblue31119180}{RGB}{31,119,180}

\begin{axis}[
tick align=outside,
tick pos=left,
x grid style={darkgray176},
xlabel={Lead Time(s)},
xmin=-0.245, xmax=5.145,
xtick style={color=black},
y grid style={darkgray176},
ylabel={Number of Data Points},
ylabel style={yshift=10pt},
ymin=-110.1, ymax=2356.1,
ytick style={color=black}
]
\addplot [draw=steelblue31119180, fill=steelblue31119180, mark=*, only marks]
table{%
x  y
0 2244
0.1 2244
0.2 2244
0.3 2244
0.4 2244
0.5 2244
0.6 2244
0.7 2241
0.8 2236
0.9 2229
1 2221
1.1 2211
1.2 2202
1.3 2107
1.4 1975
1.5 1798
1.6 1466
1.7 1182
1.8 957
1.9 785
2 654
2.1 558
2.2 493
2.3 433
2.4 399
2.5 382
2.6 346
2.7 272
2.8 231
2.9 195
3 171
3.1 152
3.2 146
3.3 118
3.4 103
3.5 87
3.6 81
3.7 78
3.8 66
3.9 51
4 33
4.1 32
4.2 32
4.3 25
4.4 21
4.5 20
4.6 17
4.7 10
4.8 7
4.9 2
};
\end{axis}

\end{tikzpicture}
    \caption{The number of data points vs. lead time.}
    \label{fig:data_points_vs_windows_data}
    \vspace{-2mm}
\end{figure}

While techniques like the Synthetic Minority Over-Sampling Technique for Regression with Gaussian Noise \cite{ref:branco2017smogn} exist for addressing imbalanced regression data, the direct correlation between lead time and false positive rate complicates achieving maximum lead time with an acceptable false positive rate, as evidenced by studies like \cite{ref:kalyanakrishnan2011learning, ref:liu}. \textbf{Consequently, we reframe the problem into a combined classification and regression challenge}. Our proposed algorithm consists of three main components: a critical fault classifier, a lead time classifier, and a lead time regressor. All three components share a 1D CNN architecture, featuring a 1D CNN layer, a max pooling layer, and two fully connected layers with the ReLu activation function. 
The critical fault classifier's objective is to predict critical faults while maximizing lead time and minimizing false positives. The lead time classifier, on the other hand, categorizes windows containing critical faults into three distinct ranges: $[0,1]$, $(1,2]$, and $(2, H]$. Notably, the $(2, H]$ range contains significantly fewer data points, as shown in Figure \ref{fig:data_points_vs_windows_data}. Finally, the lead time regressor predicts the lead time for windows that have a critical fault and a lead time within the range $[0,1]$  \footnote{The ranges of $[0,1]$, $(1,2]$, and $(2, H]$ are chosen based on the performance of the classifier and regressor.}.

\begin{figure}
\vspace{2mm}
    \centering
    \includegraphics[width=0.95\linewidth]{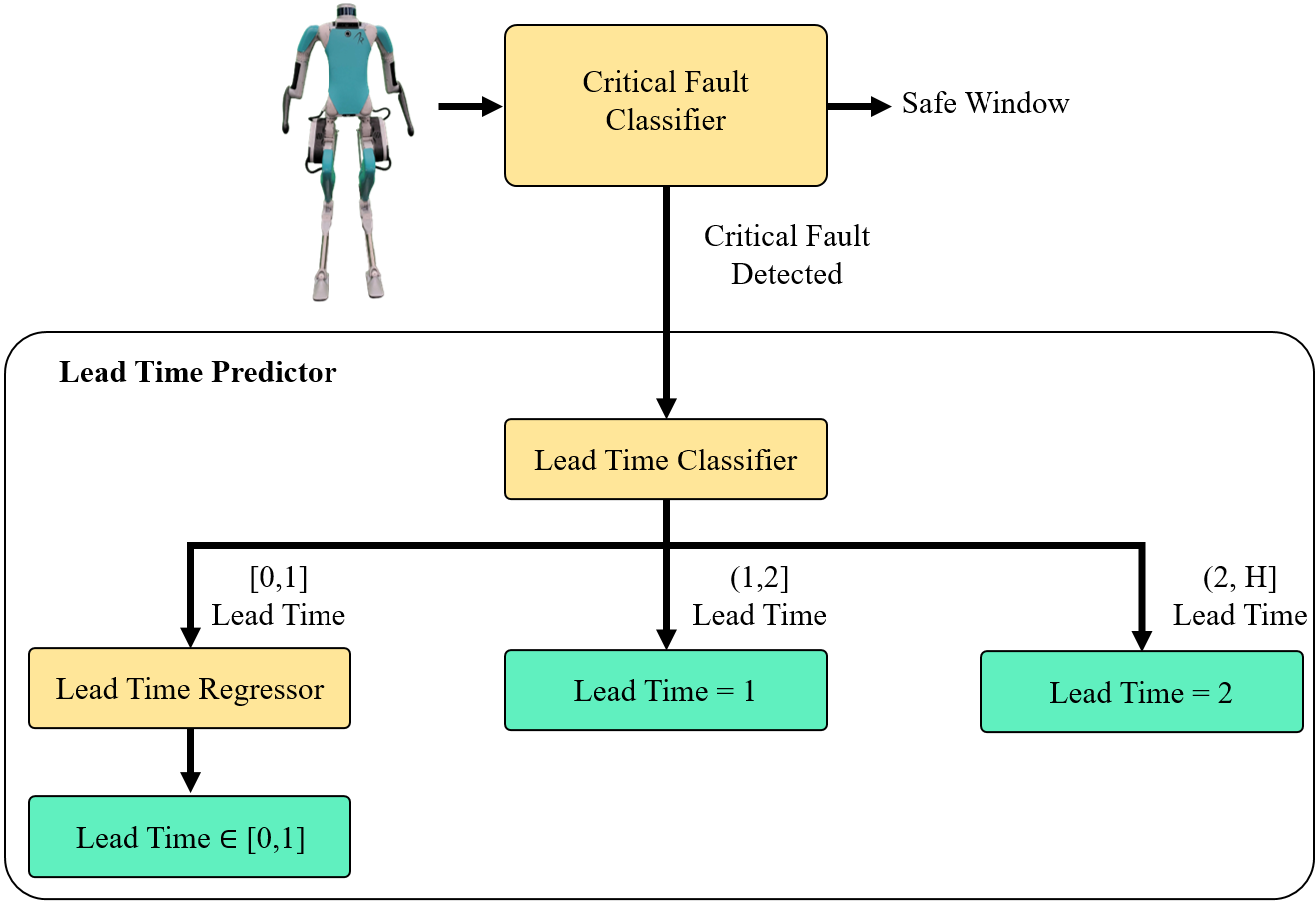} 
    \caption{The proposed fall prediction algorithm with three components: critical fault classifier, lead time classifier and lead time regressor. The green boxes depict the algorithm's predicted lead time.}
    \label{fig:fall_prediction_algorithm}
    \vspace{-2mm}
\end{figure}

The components of the algorithm interact sequentially. Initially, the critical fault classifier processes a window of data points from the robot. If a critical fault is detected, this window is then relayed to the lead time classifier, which categorizes the window based on the predefined lead time intervals. If the categorized lead time is within the $[0,1]$ interval, the lead time regressor determines the exact predicted lead time. For other intervals, the infimum lead time corresponding to that interval is reported (e.g, 1 for the interval $(1, 2]$). The entire workflow of the proposed fall prediction algorithm is depicted in Figure \ref{fig:fall_prediction_algorithm}.

\section{Critical Fault Classifier}
The features for the critical fault classifier are defined as,

\begin{equation}
\begin{bmatrix}
    (p_{com} - p_{mid toe})^{xz}\\
   v_{com}^{xz}\\
   q_{\text{sag chosen}}\\
    \Dot{q}_{\text{sag chosen}}
\end{bmatrix}
    \label{eq:features_fault_classifier}
\end{equation}
where \( p_{com} \), \( v_{com} \), and \( p_{mid toe} \) represent the position and velocity of the center of mass and the midpoint of the two toes, respectively. The terms \( q_{\text{sag chosen}} \) and \( \Dot{q}_{\text{sag chosen}} \) denote the torso, knee, hip, and toe pitch angles (in the sagittal plane).

The data for abrupt and incipient faults are structured as,
\begin{align*}
& \hspace*{3.5cm} D=\{X_i,y_i\}_{i=1}^n, \\
&~~\text{where}  \\
&\begin{array}{rcl}
    n &=& \text{number of windows across all training data}\\
    m &=& \text{number of time steps in a window}\\
    x_{ij}&=&\text{features at time step j in window i} \\    
    t_{i}&=&\text{time at time step i} \\
    t_{\text{ft}}&=&\text{time the fault is introduced} \\
    \mathbf{T}_{safe} &=&\text{safe trajectories} \\
    \mathbf{T}_{unsafe} &=&\text{unsafe trajectories} \\
        X_i &=& \begin{bmatrix} x_{i1} && x_{i2} && \cdots && x_{im}  \end{bmatrix}^\intercal \\
        \\
    y_i &\in&\  \begin{cases}
           1  & \quad  \left(X_i ~\in ~\mathbf{T}_{unsafe} \land ~ t_i~ \geq ~t_{\text{ft}} \right) \\
           \\
            0  &\quad 
\left( \begin{array}{cc}
    X_i ~\in ~ \mathbf{T}_{safe} \\
    \lor\\
      (X_i ~\in ~ \mathbf{T}_{unsafe} \land ~ t_i~ < ~t_{\text{ft}}) 
\end{array} \right). \end{cases}
    \end{array}
\end{align*}

With the above labeling, achieving correct identification for all windows would yield the maximum possible lead time. For intermittent data, the labeling approach remains similar, but \( t_{\text{ft}} \) is the time of the first fault's introduction. 

The binary cross-entropy loss is employed for training. The classifier's output, when combined with this loss, produces logits. These logits can be transformed into probabilities using the sigmoid function, allowing for flexibility in setting the desired probability threshold for critical fault detection. For instance, a model with an average of 0.27 false positive rate and 1.79s lead time for abrupt and incipient faults can achieve 0 false positive rate with 1.66s lead time by adjusting the probability threshold from 0.5 to 0.9.

The model's performance is evaluated at each epoch to optimize lead time and minimize false positive rates. A model is saved only if it meets one of the following criteria: 
\begin{enumerate}
    \item A reduction in the false positive rate of validation trajectories.
    \item No change in the validation false positive rate, but an increase in validation lead time, accompanied by a decrease in the training data's false positive rate.
    \item Both the validation and training false positive rates meet a predefined maximum threshold. 
\end{enumerate}
It's worth noting that the false positive rates used in the saving criteria pertain to entire trajectories, not individual windows.

\subsection{Training}
We train on a subset of abrupt and incipient simulation trajectories as described in Section \ref{sec:pre-processinng}. We use false positive rate, lead, and response time as our evaluation criteria. Response time is defined as the difference in time between the detection of the critical fault and its introduction. Given that the robot was not allowed to fall to the ground during hardware data collection, lead time is not defined for hardware. Similarly, given that intermittent data has two disturbances, it is difficult to estimate the response time. 

\subsection{Results}
We evaluate on hardware trajectories, as well as on the intermittent and remaining incipient and abrupt trajectories, as defined in Section \ref{sec:pre-processinng}. The critical fault classifier is able to achieve 0 false positive rate for training and validation data when trained for 3 epochs with 8 filters for the 1D CNN. When evaluated on testing data, the model is able to achieve 0 false positive and negative rates for hardware, intermittent, abrupt, and incipient fault data. 

The resulting average lead times across all unsafe trajectories are 1.64s and 1.52s for intermittent and abrupt plus incipient data, respectively. The classifier's success in categorizing the intermittent data can be attributed to the sliding window formulation, which allows the algorithm to recognize local patterns. The resulting average response times across all unsafe trajectories are 0.44s and 0.99s for the abrupt and incipient simulation data, and hardware data, respectively. The 0.55s difference in response time between the hardware and simulation data can be attributed to the lower force profiles applied in the hardware. In fact, lower force profiles in simulation resulted in similar response times to hardware.

Given that the critical fault classifier detects incipient, abrupt, and intermittent faults with a lead time greater than 0.2s, it meets our objective of detecting critical faults with sufficient lead time. The results are summarized in Table \ref{tab:classifier_results}.

\begin{table}
\vspace{2mm}
    \centering
        \caption{Results of the critical fault classifier when trained on abrupt and incipient simulation data and evaluated on (a) abrupt and incipient simulation data, (b) abrupt and incipient hardware data, and (c) intermittent data.}
    \resizebox{1.0\linewidth}{!}{
    \begin{tabular}{|c|c|c|c|c|}
    \hline
    {\bf Platform} & {\bf \makecell{Fault\\Type}} & {\bf \makecell{Average\\Lead\\Time\\ (s)}} & {\bf \makecell{Average\\Response\\Time\\ (s)} } &  {\bf \makecell{False\\Positive\\ Rate} }\\
    \hline
    \hline
       Simulation  & Intermittent & 1.64 & N/A & 0.0\\
       \hline
       Simulation &  \makecell{Abrupt and \\Incipient} & 1.52 & 0.44 & 0.0\\
       \hline
      Hardware & \makecell{Abrupt and \\Incipient}   & N/A & 0.99 & 0.0 \\
       \hline
    \end{tabular}}
    \label{tab:classifier_results}
    \vspace{-2mm}
\end{table}

\begin{figure}
\vspace{2mm}
    \centering
\captionsetup[subfigure]{labelformat=empty}
\subfloat[]{\includegraphics[scale = 0.5]{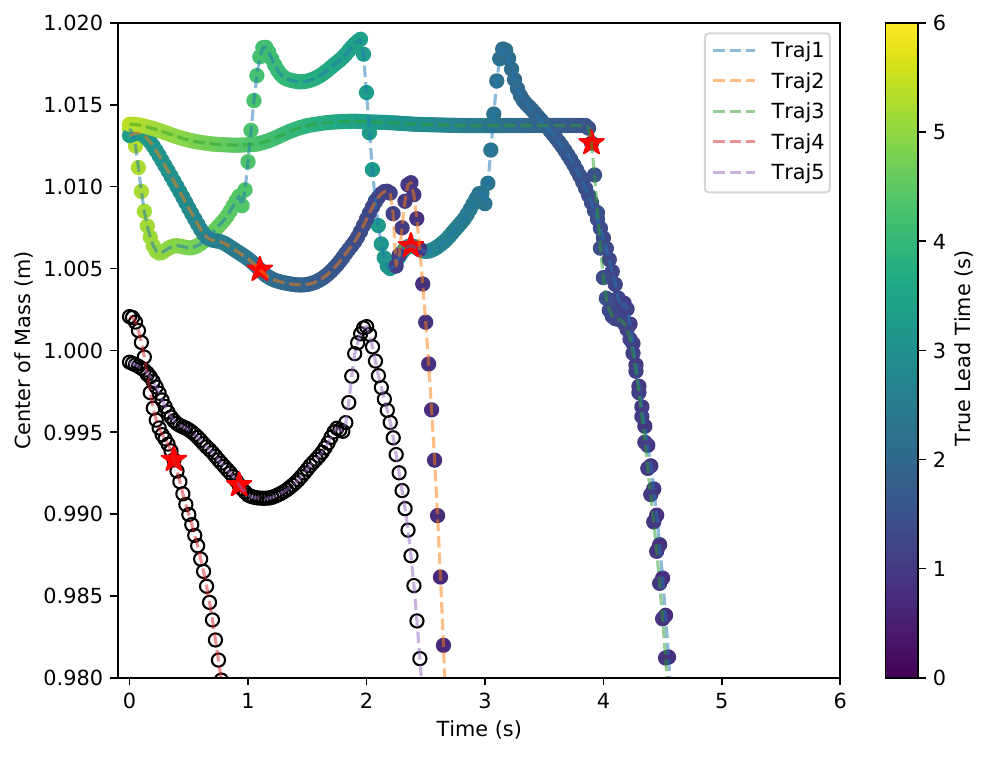}}
\captionsetup[subfigure]{labelformat=empty}
\newline\vspace{-20pt}
\subfloat[]{
\scriptsize
    \begin{tabular}{|c|c|c|c|c|c|c|}
    \hline
    & {\bf Platform} & {\bf \makecell{Fault}} & {\bf \makecell{Force(N)}} & {\bf \makecell{$\text{LT}_\text{T}$(s)} } &  {\bf \makecell{$\text{LT}_\text{P}$ (s)}} &  {\bf \makecell{RT(s)} } \\
    \hline
    \hline
       Traj1 & Sim  & AF & 224.2 & 2.95 & 2 & 2.40\\  
       \hline
       Traj2 & Sim  & IF & 31.35 & 2.25 & 2 & 1.10\\
       \hline
       Traj3 & Sim & InF & \makecell{6.32 (IF) \\ 364.2 (AF)} & 1.50 & 1 & N/A\\
       \hline
       Traj4 & Hdw & AF & N/A & N/A & 1 & 0.4\\
       \hline
       Traj5 & Hdw & IF & N/A & N/A & 1 & 0.95\\
       \hline
    \end{tabular}  
}
\caption{Performance of the algorithm in Figure \ref{fig:fall_prediction_algorithm} for several unsafe trajectories with respect to center of mass height. To conserve space, only data during and after the fault's introduction and the center of mass greater than 0.98m is included. Red stars indicate the algorithm's detection of critical faults. The dots and circles are the last data points in a window for simulation and hardware respectively. \textbf{Acronyms used:} $\text{LT}_\text{T}$ (true lead time), $\text{LT}_\text{P}$ (predicted lead time), RT (response time), Sim (simulation), Hdw (hardware), AF (abrupt fault), IF (incipient fault), and InF (intermittent fault).}
    \label{fig:traj_fall_prediction_results}
    \vspace{-2mm}
\end{figure}
\section{Lead Time Prediction}
Given the absence of a defined lead time for the hardware data, this section evaluates the lead time algorithms using only simulation data. The subsequent section will apply the complete fall prediction algorithm to both hardware and simulation datasets.

The feature set for lead time prediction includes those from \eqref{eq:features_fault_classifier}, complemented by the hip, knee, and toe pitch torques, as well as the average position of the contact point\footnote{The contact point position defaults to the rotation point when the toes rotate and to the zero moment point otherwise.}.

\subsection{Lead Time Classifier}

In this section, we detail our approach to categorizing lead times for potential falls. The categorization of lead times into three distinct intervals, $[0,1]$, $(1,2]$, and $(2,H]$, is treated as a multi-classification challenge. We employ the cross entropy loss for this purpose.

The method of data labeling is analogous to the one adopted for the critical fault classifier. However, in this context, $y_i$ is adjusted to indicate the specific lead time range to which a window is associated. A noteworthy aspect of our data is the exponential decline in the number of data points as lead time increases; recall Fig.~\ref{fig:data_points_vs_windows_data}. This implies that the amount of data available for the interval $(2,H]$ is significantly less than for the other intervals. As a result, achieving a high classification accuracy for this range is not our primary objective, but rather, an anticipated challenge.

After training on abrupt and incipient simulation faults, evaluation yields:
\begin{itemize}
    \item The classifier achieves an accuracy of $1.0$ for the interval $[0,1]$;
    \item For the interval $(1,2]$, the accuracy stands at $0.95$; and
    \item The interval $(2,H]$ sees a lower accuracy of $0.74$, in line with our expectations.
\end{itemize}

\subsection{Lead Time Regressor}

In this section, we delve into the methodology and results associated with our lead time regressor. The regressor is specifically trained on abrupt and incipient simulation data that fall within the range of $[0,1]$. For the loss function, we employ the mean squared error, which is particularly effective for regression problems as it emphasizes larger errors over smaller ones.

Upon evaluation:

\begin{itemize}
    \item The maximum difference between the predicted and actual lead times is 0.09s.
    \item The mean difference stands at 0.01s.
    \item The median difference is also 0.01s.
\end{itemize}

Given the minimal prediction error, it's evident that the lead time regressor performs well at predicting lead times within the specified range of $[0,1]$.

\section{Overall Fall Prediction Method Results}
In this section, we assess the performance of the fall prediction algorithm depicted in Figure \ref{fig:fall_prediction_algorithm} using both hardware and simulation data. It should be noted that the average lead times and false positive rates presented in Table \ref{tab:fall_prediction_results} and Figure \ref{fig:traj_fall_prediction_results} relate to the entire fall prediction algorithm, not only the critical fault classifier. 
Therefore, when the critical fault classifier identifies a critical fault within the intervals of $(1,2]$ and $(2,H]$, the true lead time is taken as 1 and 2, respectively, in line with the lead time classifier. 
When assessed on simulation data, the fall prediction algorithm predicted an average lead time of 1.18s, with a slight difference of only 0.02s from the average true lead time of 1.16s.

Next, we evaluate the algorithm on the hardware data and the simulation data trimmed at 0.95m, which is the minimum center of mass height Digit achieved during experiments. The resulting average lead time is 1.04s and 1.18s for the hardware and simulation data, respectively. Given that the fall prediction algorithm outputs the same average lead time for the simulation data at 0.95m and the original fall height of 0.12m, we can conclude that for the simulation data, the fall prediction algorithm identifies critical faults before 0.95m. In comparison to the simulation data trimmed at 0.95m and the average true lead time for simulation, the average predicted lead time for the hardware data only differs by 0.14s and 0.12s, respectively. This minute difference, compared to the average lead times greater than 1,  demonstrates the algorithm's success in identifying and predicting lead time for both hardware and simulation data. Table \ref{tab:fall_prediction_results} summarizes the results of the fall prediction algorithm, while Figure \ref{fig:traj_fall_prediction_results} displays the results for several individual trajectories.
 
\begin{table}
\vspace{1.5mm}
    \centering
    \caption{Results of the entire fall prediction algorithm when trained on abrupt and incipient simulation data and evaluated on (a) abrupt and incipient simulation data, and (b) abrupt and incipient hardware data.}
    \begin{tabular}{|c|c|c|c|c|}
    \hline
    {\bf Platform} & {\bf \makecell{Fault\\ Type}} & {\bf \makecell{Fall \\ Height\\ (m)}} & {\bf \makecell{Average\\Predicted\\Lead \\Time\\ (s)} } &  {\bf \makecell{Average\\True\\Lead\\Time\\ (s)} }\\
    \hline
    \hline
       Simulation  & \makecell{Abrupt and \\ Incipient} & 0.12 & 1.18 & 1.16\\
       \hline
       Simulation &  \makecell{Abrupt and \\Incipient} & 0.95 & 1.18 &  N/A\\
       \hline
      Hardware & \makecell{Abrupt and \\Incipient}   & 0.95 & 1.04 & N/A \\
       \hline
    \end{tabular}
    \label{tab:fall_prediction_results}
    
    \vspace{-1.5mm}
\end{table}

\section{Conclusion}
In conclusion, this paper aimed to develop an effective fall prediction algorithm for the bipedal robot Digit, considering abrupt, incipient, and intermittent faults while accurately predicting lead time. While a regression approach faces challenges due to data imbalance, we propose a comprehensive algorithm comprising a critical fault classifier, lead time classifier, and lead time regressor.

Our evaluation, based on simulation and hardware data, demonstrates the effectiveness of each component. The critical fault classifier achieves a 0 false positive rate, detecting faults with a minimum average lead time of 1.52s and a maximum average response time of 0.44s for simulation data. When evaluated on hardware data, a 0.55s time discrepancy in the average lead time was noted compared to simulation. This variance is attributed to differing force profiles in hardware. The lead time classifier exhibits an accuracy of up to 1.0 for data-rich ranges, and the lead time regressor accurately predicts lead times within a 0.09s difference in simulation data.

Assessing the entire algorithm, we observed a negligible 0.02s difference between average predicted and true lead times in simulation. In hardware evaluation, where true lead times were unknown, only a 0.12s difference in lead time was noted when compared to simulation. Given our algorithm's performance across diverse datasets and the successful prediction of fall events in bipedal robots, we can conclude that our objective has been achieved. 

After submission for review, we implemented the algorithm online on hardware and simulation. Preliminary results show successful detection of critical faults introduced by applying forces in various directions to the torso and other links.

\textbf{Acknowledgements:} Funding for M.E. Mungai was provided in part by a Rackham Merit Fellowship and in part by NSF Award No.~1808051.  Funding for J. Grizzle was provided by NSF Award No.~1808051. M.E. Mungai thanks Prof. Maani Ghaffari for useful discussions, and Grant Gibson for helpful advice and assistance in experiments.

\balance
\bibliographystyle{template/IEEEtran}
\bibliography{references.bib}
\end{document}